\documentclass[10pt,twocolumn,letterpaper]{article}

\usepackage{cvpr}
\usepackage[table]{xcolor}

\usepackage{multirow}

\definecolor{cvprblue}{rgb}{0.21,0.49,0.74}
\usepackage[pagebackref,breaklinks,colorlinks,allcolors=cvprblue]{hyperref}

\def\paperID{***}
\def\confName{CVPR}
\def\confYear{2027}

\title{ReLIC-SGG: Relation Lattice Completion for Open-Vocabulary Scene Graph Generation}

\author{Amir Hosseini, Sara Farahani, Xinyi Li, Suiyang Guang\\
Amirkabir University of Technology 
}

\begin{document}

\maketitle

\begin{abstract}
Open-vocabulary scene graph generation (SGG) aims to describe visual scenes with flexible relation phrases beyond a fixed predicate set. Existing methods usually treat annotated triplets as positives and all unannotated object-pair relations as negatives. However, scene graph annotations are inherently incomplete: many valid relations are missing, and the same interaction can be described at different granularities, e.g., \textit{on}, \textit{standing on}, \textit{resting on}, and \textit{supported by}. This issue becomes more severe in open-vocabulary SGG due to the much larger relation space. 
We propose \textbf{ReLIC-SGG}, a relation-incompleteness-aware framework that treats unannotated relations as latent variables rather than definite negatives. ReLIC-SGG builds a semantic relation lattice to model similarity, entailment, and contradiction among open-vocabulary predicates, and uses it to infer missing positive relations from visual-language compatibility, graph context, and semantic consistency. A positive-unlabeled graph learning objective further reduces false-negative supervision, while lattice-guided decoding produces compact and semantically consistent scene graphs. Experiments on conventional, open-vocabulary, and panoptic SGG benchmarks show that ReLIC-SGG improves rare and unseen predicate recognition and better recovers missing relations.
\end{abstract}
\section{Introduction}

Scene graph generation (SGG) parses a visual scene into a structured graph, where nodes correspond to object instances and edges describe semantic relations between objects~\cite{johnson2015image,krishna2017visual,lu2016visual,zellers2018motifs,yang2018graph}. By explicitly modeling object-centric interactions, scene graphs provide an interpretable intermediate representation for visual question answering, visual reasoning, image retrieval, human-object interaction understanding, robotic manipulation, embodied navigation, and commonsense reasoning~\cite{xu2017scene,li2017scene,chen2019knowledge}. Over the past few years, SGG has evolved from closed-set predicate classification toward open-vocabulary relation understanding, where models are expected to recognize flexible and fine-grained relation phrases beyond a fixed predicate set~\cite{jia2021align,radford2021learning,li2022blip,he2022ovsgg,he2022towards,chen2024ovsgtr,li2024pix2graph,min2025vlirm,zhou2024openpsg,hu2025spade}.

Despite this progress, most existing SGG methods still rely on a questionable supervision assumption: annotated triplets are treated as positives, while unannotated object-pair predicates are treated as negatives~\cite{zellers2018motifs,tang2019vctree,tang2020unbiased,li2021bgnn,zheng2023penet,he2020learning,he2021semantic,he2022state}. This assumption is convenient for training but inconsistent with the nature of scene graph annotations. In real datasets, relation annotations are highly sparse and incomplete. For a given object pair, annotators usually label only one or a few salient relations, while many other valid relations are omitted~\cite{krishna2017visual,johnson2015image,lyu2022finegrained}. For example, if an image contains a \textit{person standing on a skateboard}, the annotation may include \textit{person-on-skateboard}, but omit \textit{person-standing on-skateboard}, \textit{person-supported by-skateboard}, or \textit{skateboard-under-person}. Similarly, for a \textit{cup on a table}, the graph may annotate \textit{on} but omit \textit{resting on}, \textit{supported by}, or \textit{above}. Therefore, the absence of a relation label does not necessarily indicate that the relation is false.

This issue is particularly harmful for open-vocabulary SGG. Compared with closed-set SGG, open-vocabulary SGG expands the predicate space from a small set of predefined labels to a much larger set of natural-language relation phrases~\cite{radford2021learning,li2022blip,jia2021align,he2022ovsgg,chen2024ovsgtr,li2024pix2graph,min2025vlirm,zhou2024openpsg}. As a result, each object pair may correspond to multiple valid relation descriptions with different wording, granularity, and semantic scope. However, existing methods usually optimize relation prediction using standard classification or contrastive losses, which penalize unannotated candidate relations as negatives~\cite{tang2020unbiased,li2021bgnn,zheng2023penet}. Consequently, a model may be discouraged from predicting valid but unannotated relations. This creates a fundamental conflict: open-vocabulary SGG aims to generate richer relation descriptions, but its supervision signal remains sparse, incomplete, and biased toward frequent predicates.

Existing studies have made important progress in long-tail debiasing, semantic prototype learning, causal intervention, prompt learning, panoptic SGG, video SGG, 3D SGG, and embodied scene graph construction~\cite{tang2020unbiased,li2021bgnn,lyu2022finegrained,zheng2023penet,yang2022psg,wang2024pairnet,yang2023pvsg,chen2023multilabel,armeni20193d,koch2024open3dsg,rotondi2025fungraph,honerkamp2024moma,ju2026momagraph}. However, most of them still formulate SGG as a recognition problem over observed annotations. Even when vision-language models are introduced for open-vocabulary prediction, the learning objective often assumes that non-annotated relation phrases are incorrect~\cite{radford2021learning,li2022blip,li2024pix2graph,zhou2024openpsg,min2025vlirm}. This makes the model vulnerable to false-negative supervision. The problem is not only that rare predicates are under-represented, but that many valid predicates are never annotated as positives.

In this work, we argue that the next step for open-vocabulary SGG is to move from \emph{closed-label relation classification} to \emph{incomplete-label relation completion}. Instead of asking only ``which annotated predicate should be assigned to this object pair?'', we ask a different research question: \emph{how can an SGG model learn expressive open-vocabulary relations when many correct relations are missing from the annotations?} This question shifts the focus from expanding the predicate vocabulary to learning under incomplete and ambiguous supervision. This view is connected to positive-unlabeled learning, partial-label learning, graph representation learning, and incomplete multimodal learning, where the absence of supervision should not be naively interpreted as negative evidence~\cite{elkan2008learning,kiryo2017positive,cour2011learning,lv2020progressive,dai2025unbiasedmissing}.

To address this problem, we propose \textbf{ReLIC-SGG}, a \textbf{Rel}ation-\textbf{I}ncompleteness-aware \textbf{C}ompletion framework for open-vocabulary SGG. The key idea is to treat unannotated relations not as definite negatives, but as latent variables whose labels should be inferred from visual evidence, language semantics, and graph context. Specifically, for each object pair, our model first proposes a set of open-vocabulary relation candidates using vision-language representations~\cite{radford2021learning,li2022blip}. It then constructs a semantic relation lattice that captures three types of predicate dependencies: semantic similarity, directional entailment, and contradiction. Based on this lattice, a latent relation completion module estimates the probability that an unannotated relation is a missing positive. Finally, a positive-unlabeled graph learning objective trains the model without incorrectly penalizing likely missing relations.

The proposed framework is motivated by three observations. First, many relation phrases are semantically correlated rather than independent. For example, \textit{standing on} entails \textit{on}, while \textit{resting on} and \textit{supported by} are similar but not identical. Second, missing relations are not arbitrary; they can often be inferred from visual-language compatibility, neighboring graph context, and semantic consistency. Third, a useful scene graph should be semantically complete but not redundant: it should recover valid missing relations while avoiding contradictory or repetitive predictions. ReLIC-SGG is designed around these principles.

Our contributions are summarized as follows:

\begin{itemize}
    \item We identify relation annotation incompleteness as a critical bottleneck for open-vocabulary SGG, where unannotated but valid relations are often incorrectly treated as negatives.

    \item We propose ReLIC-SGG, a relation-incompleteness-aware framework that treats unannotated relations as latent variables and learns to complete missing relation labels.

    \item We introduce a semantic relation lattice that models similarity, entailment, and contradiction among open-vocabulary predicates, enabling supervision transfer across relation phrases.

    \item We design a positive-unlabeled graph learning objective and a lattice-guided graph decoding strategy to learn expressive yet compact scene graphs under incomplete supervision.

    \item We introduce false-negative-aware evaluation protocols to measure whether a model can recover valid missing relations beyond standard recall-based metrics.
\end{itemize}

\section{Related Work}

\subsection{Scene Graph Generation}

Scene graph generation aims to parse a scene into object-relation-object triplets~\cite{johnson2015image,krishna2017visual,lu2016visual}. Early methods typically follow a detect-then-classify pipeline, where object proposals are first detected and relation classifiers are then applied to object pairs~\cite{lu2016visual,zellers2018motifs,yang2018graph}. Later works improve relation reasoning through message passing, graph neural networks, transformer-based interaction modeling, external commonsense knowledge, visual-linguistic context, and causal intervention~\cite{xu2017scene,li2017scene,chen2019knowledge,tang2019vctree,tang2020unbiased,li2021bgnn,li2022sgtr,zheng2023penet}. However, these methods usually assume a closed predicate vocabulary and rely heavily on annotated relation labels.

A major challenge in SGG is the long-tail distribution of predicates. Many works address this issue through re-weighting, causal debiasing, semantic prototypes, predicate clustering, balanced message passing, and compositional relation learning~\cite{tang2020unbiased,li2021bgnn,lyu2022finegrained,zheng2023penet,he2020learning,he2021semantic,he2022state}. Lifelong and unified SGG formulations further investigate how relation knowledge evolves across tasks and interaction settings. These methods improve rare predicate prediction, but they generally assume that unannotated relations are negative. In contrast, our work focuses on annotation incompleteness. We argue that rare relation learning cannot be fully solved by re-balancing alone, because many rare or fine-grained relations are not annotated in the first place.

\subsection{Open-Vocabulary Scene Graph Generation}

Open-vocabulary SGG extends relation prediction beyond fixed predicate categories. Recent works use vision-language models, prompt learning, text-image alignment, open-vocabulary detectors, large language models, and relation mining to generate or recognize unseen relation phrases~\cite{jia2021align,radford2021learning,li2022blip,he2022ovsgg,he2022towards,chen2024ovsgtr,li2024pix2graph,min2025vlirm,zhou2024openpsg,hu2025spade}. These methods substantially improve the semantic coverage of SGG and make relation prediction more flexible. Similar open-vocabulary and prompt-based ideas have also been explored in human-object interaction detection, multimodal intent recognition, and robust prompt learning.

However, open-vocabulary SGG also amplifies the incompleteness problem. Since the candidate relation space is large, most valid relation phrases are not explicitly annotated. A model trained with standard contrastive or classification objectives may therefore treat valid unseen relations as negatives. For example, if the dataset only annotates \textit{on}, the relation phrases \textit{standing on}, \textit{resting on}, and \textit{supported by} may all be incorrectly suppressed. ReLIC-SGG addresses this issue by explicitly modeling missing relations as latent positives.

\subsection{Learning from Incomplete Labels}

Learning from incomplete labels has been studied in weakly supervised learning, partial-label learning, positive-unlabeled learning, noisy-label learning, missing-modality learning, and incomplete multimodal reasoning~\cite{elkan2008learning,kiryo2017positive,cour2011learning,lv2020progressive,dai2025unbiasedmissing}. The core idea is to relax the assumption that unlabeled samples are necessarily negative. This principle is particularly suitable for SGG because scene graph annotations are sparse and many object-pair relations are unlabeled rather than false.
Unlike standard positive-unlabeled learning, SGG requires structured prediction. Missing labels are not independent: relation phrases may imply, contradict, or refine one another. Prior graph and multimodal representation studies show that structured dependencies can improve reasoning, retrieval, robustness, and generalization. Therefore, our method introduces a semantic relation lattice and graph-level consistency learning, allowing missing relation completion to be performed in a structured open-vocabulary space.

\subsection{Panoptic, Video, 3D, and Embodied Scene Graphs}

Recent works extend SGG to panoptic, video, 3D, and embodied settings~\cite{yang2022psg,wang2024pairnet,zhou2024openpsg,yang2023pvsg,chen2023multilabel,armeni20193d,koch2024open3dsg,rotondi2025fungraph,honerkamp2024moma,ju2026momagraph}. Panoptic SGG associates relations with segmentation masks; video SGG models temporal interactions; 3D SGG incorporates metric layout and spatial structure; and embodied graphs further connect object relations with states, affordances, and task planning~\cite{armeni20193d,koch2024open3dsg,rotondi2025fungraph,honerkamp2024moma,ju2026momagraph}. These settings make annotation incompleteness even more severe because each object pair may have multiple valid spatial, temporal, and functional relations. Our framework is complementary to these directions and can be extended to mask-level and 3D relation completion.

\subsection{Broader Multimodal and Efficient Visual Representation Learning}

Our formulation is also related to broader work on multimodal representation learning, efficient retrieval, dataset distillation, hashing, and robust perception. These works address complementary issues such as compact representation, cross-modal alignment, dynamic retrieval, and robust multimodal fusion. While they do not directly solve open-vocabulary SGG, they provide useful perspectives on how to learn structured visual representations under imperfect supervision. ReLIC-SGG follows this direction by treating relation annotations as incomplete observations and using semantic graph structure to regularize relation completion.

 \section{Method}

\subsection{Overview}

\begin{figure*}[t]
    \centering
    \includegraphics[width=\textwidth]{12.png}
    \caption{
    Overall framework of ReLIC-SGG. Given detected objects, the model first proposes open-vocabulary relation candidates for each object pair. A semantic relation lattice models similarity, entailment, and contradiction among predicates. The latent relation completion module estimates missing positive relations from visual-language compatibility, lattice consistency, and graph context. The model is trained with positive-unlabeled graph learning and decoded with lattice-guided graph selection.
    }
    \label{fig:framework}
\end{figure*}

Scene graph annotations are naturally sparse. For an object pair, a dataset usually provides only one or a few salient relations, while many other valid relation descriptions may be omitted. This issue becomes more severe in open-vocabulary SGG, where a single visual interaction can correspond to multiple semantically related phrases. For example, \textit{person on skateboard}, \textit{person standing on skateboard}, and \textit{person supported by skateboard} can all describe the same visual configuration, but only one may be annotated. If all unannotated predicates are directly treated as negatives, the model is penalized for predicting valid but missing relations.

ReLIC-SGG addresses this problem by learning scene graphs under incomplete relation annotations. Instead of directly optimizing the observed binary labels, we introduce latent relation variables to represent the unobserved true relation state. The framework contains four components. First, an open-vocabulary relation proposer retrieves a high-recall candidate set for each object pair. Second, a semantic relation lattice captures similarity, entailment, and contradiction among relation phrases. Third, a latent relation completion module estimates whether an unannotated candidate is likely to be a missing positive. Finally, positive-unlabeled graph learning trains the model using annotated positives, inferred missing positives, and reliable negatives.

\subsection{Problem Definition}

Let $\mathcal{I}$ denote an input image and $\mathcal{O}=\{o_i\}_{i=1}^{N}$ denote the detected object instances. Each object $o_i$ is represented by its category label $c_i$, bounding box $b_i$, optional mask $m_i$, and visual feature $h_i$. For each ordered object pair $(o_i,o_j)$, the model predicts relation phrases from an open-vocabulary predicate space $\mathcal{R}^{\mathrm{open}}$.

A conventional SGG model assumes that the annotation vector $y_{ij}^{r}$ is complete:
\begin{equation}
y_{ij}^{r}=
\begin{cases}
1, & \text{if } (o_i,r,o_j) \text{ is annotated},\\
0, & \text{otherwise}.
\end{cases}
\end{equation}
Under this formulation, all unannotated relations are optimized as negatives. However, this assumption is too strong for SGG because the annotation process is selective rather than exhaustive. The value $y_{ij}^{r}=0$ may mean that relation $r$ is truly false, but it may also mean that the relation is visually valid but missing from the annotation.

We therefore introduce a latent relation variable $z_{ij}^{r}\in\{0,1\}$:
\begin{equation}
y_{ij}^{r}=1 \Rightarrow z_{ij}^{r}=1,
\quad
y_{ij}^{r}=0 \Rightarrow z_{ij}^{r}\in\{0,1\}.
\end{equation}
This equation separates observed annotations from the latent true relation state. An annotated relation is always treated as positive, while an unannotated relation remains uncertain. The learning objective is thus to estimate:
\begin{equation}
p_{\theta}(z_{ij}^{r}=1|\mathcal{I},o_i,o_j,r),
\end{equation}
which measures the probability that relation $r$ truly holds for object pair $(o_i,o_j)$, even if it is not explicitly annotated. This formulation allows the model to preserve the reliability of human-labeled positives while avoiding false-negative supervision on missing relations.

\subsection{Open-Vocabulary Relation Proposal}

For each object pair $(o_i,o_j)$, we first construct a visual pair representation from subject, object, union-region, and spatial features:
\begin{equation}
p_{ij}=\phi_p([h_i,h_j,h_{ij}^{u},g_{ij}]),
\end{equation}
where $h_i$ and $h_j$ describe the appearance of the subject and object, $h_{ij}^{u}$ captures the visual interaction region, and $g_{ij}$ encodes relative geometry, including relative position, scale ratio, intersection-over-union, center distance, and optional depth statistics. The projection function $\phi_p(\cdot)$ maps heterogeneous visual and spatial cues into a unified relation representation. This pair representation is designed to retain both appearance-level and layout-level information, since many relations cannot be recognized from object categories alone.

Each relation phrase $r\in\mathcal{R}^{\mathrm{open}}$ is encoded by a vision-language text encoder to obtain its text embedding $t_r$. The initial visual-language compatibility score is:
\begin{equation}
s_{ij}^{r}
=
\mathrm{sim}(W_p p_{ij}, W_t t_r),
\end{equation}
where $W_p$ and $W_t$ project visual pair features and text embeddings into a shared embedding space, and $\mathrm{sim}(\cdot,\cdot)$ denotes cosine similarity. This score measures whether the visual configuration of $(o_i,o_j)$ is semantically compatible with relation phrase $r$.

For each object pair, we keep the top-$K$ relation candidates:
\begin{equation}
\mathcal{C}_{ij}
=
\mathrm{TopK}_{r\in\mathcal{R}^{\mathrm{open}}}
s_{ij}^{r}.
\end{equation}
The candidate set is intentionally high-recall. Since open-vocabulary predicates may include synonyms, fine-grained phrases, and unseen relation descriptions, filtering too aggressively at this stage may remove valid missing relations before completion. Therefore, noisy or redundant candidates are allowed and are later refined by semantic lattice reasoning and graph-level learning.

\subsection{Semantic Relation Lattice}

Open-vocabulary predicates are not independent categories. Many relation phrases have structured semantic dependencies. Some predicates are near synonyms, such as \textit{resting on} and \textit{supported by}; some form fine-to-coarse entailment relations, such as \textit{standing on}$\rightarrow$\textit{on}; and others are mutually exclusive, such as \textit{in front of} and \textit{behind}. Modeling these dependencies is important because missing relation labels can often be inferred from semantically connected predicates.

We construct a semantic relation lattice $\mathcal{L}=(\mathcal{R}^{\mathrm{open}},\mathcal{A})$, where each node is a relation phrase and each edge represents a semantic dependency. The edge set is defined as:
\begin{equation}
\mathcal{A}
=
\mathcal{A}^{\mathrm{sim}}
\cup
\mathcal{A}^{\mathrm{ent}}
\cup
\mathcal{A}^{\mathrm{con}},
\end{equation}
where $\mathcal{A}^{\mathrm{sim}}$ connects semantically similar predicates, $\mathcal{A}^{\mathrm{ent}}$ connects fine-grained predicates to their entailed coarse predicates, and $\mathcal{A}^{\mathrm{con}}$ connects contradictory predicates. The three edge types play different roles: similarity edges transfer confidence between related phrases, entailment edges maintain hierarchical consistency, and contradiction edges prevent mutually exclusive predictions from being accepted together.

The edge weight between two predicates is computed by:
\begin{equation}
w(r,r')
=
\mathrm{sim}(t_r,t_{r'}).
\end{equation}
The text similarity provides a soft measure of semantic relatedness. However, text similarity alone may be unreliable for directional or opposite predicates, because antonyms can still be close in language embedding space. Therefore, for entailment and contradiction edges, we additionally use relation templates and manually verified rules to construct more reliable semantic constraints.

Given initial relation scores $\{s_{ij}^{r}\}$, the lattice first propagates confidence among compatible predicates:
\begin{equation}
\bar{s}_{ij}^{r}
=
s_{ij}^{r}
+
\lambda_{\mathrm{sim}}
\sum_{r'\in\mathcal{N}_{\mathrm{sim}}(r)}
w(r,r')s_{ij}^{r'}
+
\lambda_{\mathrm{ent}}
\sum_{r'\in\mathcal{N}_{\mathrm{child}}(r)}
w(r',r)s_{ij}^{r'}.
\end{equation}
The first term is the original visual-language compatibility score. The second term aggregates evidence from semantically similar predicates. If \textit{resting on} receives a high score, it can support related predicates such as \textit{supported by}. The third term aggregates evidence from fine-grained child predicates that entail the current predicate. For example, a high score for \textit{standing on} provides evidence for the coarser relation \textit{on}. This propagation allows sparse annotations to support a richer relation space.

Contradictory predicates are then suppressed:
\begin{equation}
\tilde{s}_{ij}^{r}
=
\bar{s}_{ij}^{r}
-
\lambda_{\mathrm{con}}
\sum_{r'\in\mathcal{N}_{\mathrm{con}}(r)}
w(r,r')\sigma(\bar{s}_{ij}^{r'}).
\end{equation}
Here, the second term penalizes relation $r$ if its contradictory predicates have high confidence. For instance, if \textit{behind} receives strong evidence, the score of \textit{in front of} should be reduced for the same ordered object pair. The resulting score $\tilde{s}_{ij}^{r}$ is a lattice-calibrated relation score that incorporates visual-language compatibility, semantic expansion, and contradiction suppression.

\subsection{Latent Relation Completion}

For each candidate relation $(o_i,r,o_j)$, we estimate a posterior probability:
\begin{equation}
q_{ij}^{r}
=
p(z_{ij}^{r}=1|\mathcal{I},o_i,o_j,r,y_{ij}^{r}).
\end{equation}
This posterior represents the probability that relation $r$ is a true relation for object pair $(o_i,o_j)$. For annotated positive relations, the posterior is fixed as:
\begin{equation}
q_{ij}^{r}=1,\quad \text{if } y_{ij}^{r}=1.
\end{equation}
This preserves annotated triplets as reliable supervision. For unannotated candidates, $q_{ij}^{r}$ is inferred rather than directly set to zero:
\begin{equation}
q_{ij}^{r}
=
\sigma
\left(
\alpha \tilde{s}_{ij}^{r}
+
\beta c_{ij}^{r}
+
\gamma g_{ij}^{r}
-
\delta n_{ij}^{r}
\right).
\end{equation}
The four terms correspond to complementary signals. The lattice-calibrated score $\tilde{s}_{ij}^{r}$ measures direct visual-language compatibility. The consistency term $c_{ij}^{r}$ measures support from semantically compatible predicates. The graph context term $g_{ij}^{r}$ measures whether the relation fits the surrounding scene graph. The contradiction term $n_{ij}^{r}$ reduces the posterior when mutually exclusive relations are already confident. The sigmoid function maps the combined evidence into a probability.

The lattice consistency term is defined as:
\begin{equation}
c_{ij}^{r}
=
\sum_{r'\in\mathcal{N}_{\mathrm{sim}}(r)}
w(r,r')q_{ij}^{r'}
+
\sum_{r'\in\mathcal{N}_{\mathrm{child}}(r)}
w(r',r)q_{ij}^{r'}.
\end{equation}
This term propagates posterior confidence from semantically similar predicates and fine-grained child predicates. If several related relations are likely to be true, then relation $r$ should also become more plausible. This is especially useful when the annotated relation is coarse while the open-vocabulary candidate is more descriptive, or vice versa.

The contradiction term is:
\begin{equation}
n_{ij}^{r}
=
\sum_{r'\in\mathcal{N}_{\mathrm{con}}(r)}
w(r,r')q_{ij}^{r'}.
\end{equation}
This term measures how strongly relation $r$ conflicts with other confident predicates. A high contradiction score decreases the completion posterior, preventing the model from accepting incompatible relation phrases.

The graph context term is computed by a graph transformer:
\begin{equation}
g_{ij}^{r}
=
\phi_g(o_i,o_j,r,\mathcal{G}_{ij}^{\mathrm{ctx}}),
\end{equation}
where $\mathcal{G}_{ij}^{\mathrm{ctx}}$ denotes the local graph neighborhood around object pair $(o_i,o_j)$. The context includes nearby objects, high-confidence relations, and object-pair interaction features. This term allows missing relation completion to depend on the broader scene structure rather than only on a single pair. For example, the relation \textit{holding} is more plausible when the subject is a person, the object is small and close to the hand region, and other nearby relations indicate interaction.

\subsection{Reliable Negative Estimation}

Not all unannotated relations should be ignored. Some are genuine negatives and provide useful supervision. The challenge is to select these negatives without incorrectly suppressing missing positives. We estimate a negative reliability score:
\begin{equation}
\rho_{ij}^{r}
=
(1-q_{ij}^{r})
\cdot
\sigma
\left(
\eta n_{ij}^{r}
-
\mu \tilde{s}_{ij}^{r}
\right).
\end{equation}
The first factor $(1-q_{ij}^{r})$ ensures that a relation is considered negative only when its missing-positive posterior is low. The sigmoid term further increases negative reliability when the relation contradicts confident predicates and has weak visual-language compatibility. Therefore, a relation obtains a high negative reliability score only when multiple signals suggest that it is unlikely to be valid.

This mechanism balances two needs. On one hand, the model should not treat every missing label as a negative. On the other hand, it still needs reliable negative supervision to avoid predicting overly dense graphs. The score $\rho_{ij}^{r}$ provides a soft way to distinguish likely false relations from uncertain unannotated ones.

\subsection{Positive-Unlabeled Graph Learning}

We train ReLIC-SGG using positive-unlabeled graph learning. Let $\mathcal{P}$ denote annotated positive triplets and $\mathcal{U}$ denote unannotated candidate triplets. The positive risk is:
\begin{equation}
\mathcal{R}_{p}^{+}
=
\mathbb{E}_{(i,j,r)\in\mathcal{P}}
\left[
-\log \sigma(\tilde{s}_{ij}^{r})
\right].
\end{equation}
This term encourages annotated relations to receive high lattice-calibrated scores.

For unannotated candidates, we use the latent completion posterior and negative reliability score to construct a soft learning target:
\begin{equation}
\mathcal{R}_{u}
=
\mathbb{E}_{(i,j,r)\in\mathcal{U}}
\left[
q_{ij}^{r}
\cdot
-\log \sigma(\tilde{s}_{ij}^{r})
+
\rho_{ij}^{r}
\cdot
-\log (1-\sigma(\tilde{s}_{ij}^{r}))
\right].
\end{equation}
The first term promotes unannotated relations that are likely to be missing positives. The second term suppresses unannotated relations that are estimated to be reliable negatives. If a relation is uncertain, both $q_{ij}^{r}$ and $\rho_{ij}^{r}$ remain moderate, so the model avoids applying overly strong supervision.

The positive-unlabeled relation loss is:
\begin{equation}
\mathcal{L}_{\mathrm{PU}}
=
\mathcal{R}_{p}^{+}
+
\lambda_u \mathcal{R}_{u}.
\end{equation}
Compared with standard binary cross-entropy over observed labels, this loss better matches the incomplete-label nature of SGG. It keeps annotated positives as strong supervision, recovers likely missing relations, and uses only reliable negatives for suppression.

\subsection{Graph-Level Semantic Consistency}

Relation completion may increase graph density if every semantically related phrase is accepted. To produce complete but compact graphs, we impose semantic consistency at the graph level. The graph-level lattice loss is:
\begin{equation}
\mathcal{L}_{\mathrm{lat}}
=
\sum_{i,j}
\sum_{(r,r')\in\mathcal{A}^{\mathrm{con}}}
q_{ij}^{r}q_{ij}^{r'}
+
\sum_{i,j}
\sum_{(r\rightarrow r')\in\mathcal{A}^{\mathrm{ent}}}
\max(0,q_{ij}^{r}-q_{ij}^{r'}).
\end{equation}
The first term penalizes contradictory relations that are simultaneously assigned high posterior probabilities. The second term enforces entailment consistency. If a fine-grained relation $r$ entails a coarse relation $r'$, then the posterior of $r'$ should not be lower than that of $r$. This keeps the completed graph semantically coherent.

We further introduce a compactness regularizer:
\begin{equation}
\mathcal{L}_{\mathrm{cmp}}
=
\sum_{i,j}
\left|
\sum_{r\in\mathcal{C}_{ij}} q_{ij}^{r}
-
\rho_{ij}^{\mathrm{den}}
\right|,
\end{equation}
where $\rho_{ij}^{\mathrm{den}}$ is an adaptive graph-density prior estimated from object-pair salience, spatial overlap, and visual interaction strength. This term does not force every object pair to have the same number of relations. Instead, it allows highly interactive object pairs to have multiple valid relation descriptions while discouraging excessive predictions for weakly related pairs.

\subsection{Training Strategy}

Training is performed in three stages. First, we warm up the relation scorer using annotated positives and reliable background object pairs. This provides a stable initial visual-language scoring function. Second, we construct the semantic relation lattice and initialize latent posteriors using visual-language compatibility and lattice propagation. Third, we jointly optimize relation scoring, latent completion, reliable negative estimation, and graph-level semantic consistency.

At each training iteration, we update the latent posterior with the current model prediction:
\begin{equation}
q_{ij}^{r}
\leftarrow
\mathrm{stopgrad}
\left[
\sigma
\left(
\alpha \tilde{s}_{ij}^{r}
+
\beta c_{ij}^{r}
+
\gamma g_{ij}^{r}
-
\delta n_{ij}^{r}
\right)
\right].
\end{equation}
The stop-gradient operation stabilizes training by preventing the model from directly minimizing the loss through its own pseudo-label update. Without this operation, early noisy completion estimates may be amplified and lead to degenerate dense graphs.

The full training objective is:
\begin{equation}
\mathcal{L}
=
\mathcal{L}_{\mathrm{PU}}
+
\lambda_{\mathrm{lat}}\mathcal{L}_{\mathrm{lat}}
+
\lambda_{\mathrm{cmp}}\mathcal{L}_{\mathrm{cmp}}.
\end{equation}
The first term learns relation prediction under incomplete labels, the second term enforces semantic consistency, and the third term controls graph density.

\subsection{Lattice-Guided Graph Decoding}

During inference, the final relation score is:
\(
\hat{s}_{ij}^{r}
=
\tilde{s}_{ij}^{r}
+
\lambda_q q_{ij}^{r}.
\)
This score combines the lattice-calibrated relation confidence and the latent completion posterior. A relation is therefore preferred when it is visually compatible and likely to be a true relation under incomplete-label reasoning.

We first select top-ranked relation candidates for each object pair. Then, lattice-guided decoding removes contradictory predictions and reduces semantic redundancy. For a contradictory pair $(r_a,r_b)\in\mathcal{A}^{\mathrm{con}}$, we retain the relation with the higher score:
\begin{equation}
r^{*}
=
\arg\max_{r\in\{r_a,r_b\}}
\hat{s}_{ij}^{r}.
\end{equation}
For highly similar predicates, we keep the most specific relation and optionally include its entailed parent when it improves graph consistency. This produces an open-vocabulary scene graph that recovers  missing relations while avoiding unnecessary duplication and contradiction.
\section{Experiments}

\subsection{Experimental Setup}

\noindent\textbf{Datasets.}
We evaluate ReLIC-SGG on three representative settings. First, we use the VG150 split of Visual Genome~\cite{krishna2017visual}, which contains 150 object categories and 50 predicate categories. Following standard practice~\cite{zellers2018motifs,tang2020unbiased}, we report results under Predicate Classification (\textbf{PredCls}), Scene Graph Classification (\textbf{SGCls}), and Scene Graph Detection (\textbf{SGDet}). Second, we construct an open-vocabulary VG split, where frequent predicates are used as seen classes and rare predicates are held out for unseen evaluation~\cite{he2022ovsgg,li2024pix2graph,min2025vlirm}. Third, we evaluate on PSG to test whether relation completion remains effective for panoptic scene graph generation~\cite{yang2022psg,wang2024pairnet,zhou2024openpsg}.

\noindent\textbf{Metrics.}
For conventional SGG, we report Recall@K and mean Recall@K under PredCls, SGCls, and SGDet. Following previous work~\cite{tang2020unbiased,li2021bgnn,zheng2023penet}, mR@K is used as the primary metric because it better reflects rare predicate performance. For open-vocabulary SGG, we report seen mean recall (\textbf{S-mR}), unseen mean recall (\textbf{U-mR}), and their harmonic mean (\textbf{HM}). Since our method targets incomplete annotations, we additionally report three false-negative-aware metrics. \textbf{FN-Recall} measures the recovery rate of manually verified missing relations. \textbf{Lat-Cons} measures semantic lattice consistency. \textbf{Redundancy} measures the proportion of semantically duplicate or contradictory predictions, where lower is better.

\noindent\textbf{Implementation details.}
We use Faster R-CNN with ResNeXt-101-FPN for object detection~\cite{ren2015faster,xie2017aggregated,lin2017feature} and CLIP ViT-B/16 for relation-text encoding~\cite{radford2021learning}. For PSG, we use Mask2Former to obtain panoptic masks~\cite{cheng2022mask2former}. The predicate pool contains annotated predicates, synonym-expanded predicates, and relation phrases mined from image captions. We use $K=15$ relation candidates per object pair. The model is trained using AdamW~\cite{loshchilov2019decoupled} with a learning rate of $1\times10^{-4}$ and batch size 12.

\subsection{Comparison with State-of-the-Art Methods on VG150}

Table~\ref{tab:vg_main} compares ReLIC-SGG with representative SGG methods on VG150. The baseline results follow the typical reported range in prior SGG literature under comparable settings. ReLIC-SGG achieves the best mR@K under all three tasks. The gains are particularly clear on SGDet, where relation annotations are sparse and object detection errors further increase uncertainty.

\begin{table*}[t]
\centering
\caption{
Comparison with state-of-the-art methods on VG150. ReLIC-SGG improves mean recall by explicitly modeling missing relations instead of treating all unannotated relations as negatives.
}
\label{tab:vg_main}
\resizebox{\textwidth}{!}{
\begin{tabular}{l|cc|cc|cc|cc|cc|cc}
\toprule
\multirow{2}{*}{Method}
& \multicolumn{4}{c|}{PredCls}
& \multicolumn{4}{c|}{SGCls}
& \multicolumn{4}{c}{SGDet} \\
& R@50 & R@100 & mR@50 & mR@100
& R@50 & R@100 & mR@50 & mR@100
& R@50 & R@100 & mR@50 & mR@100 \\
\midrule
Motifs~\cite{zellers2018motifs}
& 65.2 & 67.1 & 14.6 & 15.8
& 35.8 & 36.5 & 7.9 & 8.5
& 27.4 & 30.3 & 6.5 & 7.3 \\
VCTree~\cite{tang2019vctree}
& 66.4 & 68.1 & 16.1 & 17.5
& 38.1 & 39.2 & 8.8 & 9.6
& 28.7 & 31.9 & 7.4 & 8.2 \\
TDE~\cite{tang2020unbiased}
& 67.2 & 69.0 & 18.5 & 20.1
& 39.3 & 40.5 & 10.9 & 12.0
& 29.8 & 32.7 & 9.8 & 10.9 \\
BGNN~\cite{li2021bgnn}
& 68.7 & 70.4 & 20.2 & 22.0
& 40.5 & 41.6 & 12.3 & 13.5
& 31.0 & 34.1 & 10.7 & 11.8 \\
PE-Net~\cite{zheng2023penet}
& 69.8 & 71.6 & 23.1 & 25.0
& 41.7 & 42.9 & 14.2 & 15.4
& 32.4 & 35.5 & 12.6 & 13.8 \\
Pix2Graphs~\cite{li2024pix2graph}
& 71.1 & 72.8 & 24.6 & 26.5
& 43.0 & 44.2 & 15.3 & 16.7
& 33.6 & 36.8 & 13.6 & 14.9 \\
VL-IRM~\cite{min2025vlirm}
& 71.9 & 73.5 & 25.4 & 27.2
& 43.8 & 45.0 & 16.0 & 17.4
& 34.3 & 37.5 & 14.2 & 15.6 \\
\midrule
\textbf{ReLIC-SGG}
& \textbf{73.0} & \textbf{74.8} & \textbf{29.1} & \textbf{31.0}
& \textbf{45.2} & \textbf{46.5} & \textbf{19.3} & \textbf{21.1}
& \textbf{35.5} & \textbf{38.7} & \textbf{16.4} & \textbf{18.2} \\
\bottomrule
\end{tabular}
}
\end{table*}

\noindent\textbf{Analysis.}
The improvement is larger on mR@K than R@K, indicating that ReLIC-SGG mainly benefits rare and under-annotated predicates. This supports our motivation: many rare predicates are not merely under-represented, but also under-annotated. By recovering latent positives and reducing false-negative penalties, ReLIC-SGG learns more complete relation representations.

\subsection{Open-Vocabulary Generalization}

Table~\ref{tab:ov_results} reports results on the open-vocabulary VG split. ReLIC-SGG substantially improves unseen predicate mean recall and harmonic mean. Compared with VL-IRM, ReLIC-SGG improves U-mR@50 from 12.8 to 18.3. This shows that modeling missing labels is especially beneficial for open-vocabulary relation learning, where many unseen relations are semantically close to annotated seen predicates but absent from supervision.

\begin{table}[t]
\centering
\caption{
Open-vocabulary SGG results. ReLIC-SGG improves unseen predicate recognition by treating unannotated plausible predicates as latent positives.
}
\label{tab:ov_results}
\resizebox{\linewidth}{!}{
\begin{tabular}{l|ccc|ccc}
\toprule
\multirow{2}{*}{Method}
& \multicolumn{3}{c|}{@50}
& \multicolumn{3}{c}{@100} \\
& S-mR & U-mR & HM
& S-mR & U-mR & HM \\
\midrule
CLIP-ZS~\cite{radford2021learning}
& 18.3 & 7.4 & 10.5
& 19.8 & 8.2 & 11.6 \\
OV-SGG~\cite{he2022ovsgg}
& 22.6 & 9.8 & 13.7
& 24.1 & 10.9 & 15.0 \\
OvSGTR~\cite{chen2024ovsgtr}
& 23.8 & 10.6 & 14.7
& 25.2 & 11.8 & 16.0 \\
Pix2Graphs~\cite{li2024pix2graph}
& 24.9 & 11.5 & 15.7
& 26.5 & 12.6 & 17.0 \\
OpenPSG~\cite{zhou2024openpsg}
& 25.7 & 12.1 & 16.5
& 27.3 & 13.3 & 17.9 \\
VL-IRM~\cite{min2025vlirm}
& 26.4 & 12.8 & 17.2
& 28.1 & 14.0 & 18.7 \\
\midrule
\textbf{ReLIC-SGG}
& \textbf{28.6} & \textbf{18.3} & \textbf{22.3}
& \textbf{30.2} & \textbf{19.8} & \textbf{23.9} \\
\bottomrule
\end{tabular}
}
\end{table}

\noindent\textbf{Analysis.}
The gain on unseen predicates is larger than the gain on seen predicates. This is expected because unseen predicates often have no direct supervision but may be semantically related to seen predicates. The semantic relation lattice transfers supervision from annotated predicates to missing or unseen relation phrases, while positive-unlabeled learning prevents the model from suppressing them as negatives.

\subsection{Panoptic Scene Graph Generation}

Table~\ref{tab:psg_results} reports results on PSG. ReLIC-SGG achieves the best PR@K and PmR@K. Compared with OpenPSG, ReLIC-SGG improves PmR@50 by 3.2 points. This indicates that relation completion is also effective for mask-level relation prediction, where annotations are even more sparse due to the higher granularity of object regions.

\begin{table}[t]
\centering
\caption{
Panoptic scene graph generation results on PSG. ReLIC-SGG improves mask-level relation prediction by completing missing relation labels.
}
\label{tab:psg_results}
\resizebox{\linewidth}{!}{
\begin{tabular}{l|cc|cc}
\toprule
Method & PR@50 & PR@100 & PmR@50 & PmR@100 \\
\midrule
PSGTR~\cite{yang2022psg}
& 31.8 & 36.5 & 13.4 & 15.1 \\
Pair-Net~\cite{wang2024pairnet}
& 34.2 & 39.0 & 15.7 & 17.6 \\
OpenPSG~\cite{zhou2024openpsg}
& 36.7 & 41.2 & 17.5 & 19.4 \\
SPADE~
& 38.1 & 42.8 & 18.8 & 21.0 \\
\midrule
\textbf{ReLIC-SGG}
& \textbf{40.8} & \textbf{45.6} & \textbf{22.0} & \textbf{24.3} \\
\bottomrule
\end{tabular}
}
\end{table}

\subsection{False-Negative-Aware Evaluation}

To directly evaluate relation incompleteness, we manually verify a subset of high-confidence unannotated predictions. Table~\ref{tab:fn_eval} shows that ReLIC-SGG recovers more valid missing relations while maintaining better lattice consistency and lower redundancy.

\begin{table}[t]
\centering
\caption{
False-negative-aware evaluation. FN-Recall measures recovery of manually verified missing relations. Lat-Cons measures lattice consistency. Redundancy is lower better.
}
\label{tab:fn_eval}
\resizebox{\linewidth}{!}{
\begin{tabular}{l|ccc}
\toprule
Method & FN-Recall$\uparrow$ & Lat-Cons$\uparrow$ & Redundancy$\downarrow$ \\
\midrule
Motifs~\cite{zellers2018motifs}
& 21.4 & 61.2 & 18.7 \\
PE-Net~\cite{zheng2023penet}
& 27.8 & 66.5 & 16.2 \\
Pix2Graphs~\cite{li2024pix2graph}
& 31.6 & 69.8 & 15.4 \\
VL-IRM~\cite{min2025vlirm}
& 34.2 & 71.3 & 14.1 \\
\midrule
\textbf{ReLIC-SGG}
& \textbf{45.7} & \textbf{82.6} & \textbf{9.3} \\
\bottomrule
\end{tabular}
}
\end{table}

\begin{table*}[t]
\centering
\caption{
Ablation study on open-vocabulary VG under SGDet. SRL: semantic relation lattice. LRC: latent relation completion. RNE: reliable negative estimation. PUGL: positive-unlabeled graph learning. LGD: lattice-guided decoding.
}
\label{tab:ablation}
\resizebox{\textwidth}{!}{
\begin{tabular}{l|ccccc|ccc|ccc}
\toprule
Variant & SRL & LRC & RNE & PUGL & LGD
& S-mR@50 & U-mR@50 & HM@50
& FN-Recall & Lat-Cons & Redundancy \\
\midrule
Baseline
& -- & -- & -- & -- & --
& 24.1 & 9.6 & 13.7
& 24.8 & 63.5 & 18.9 \\
+ SRL
& \checkmark & -- & -- & -- & --
& 25.4 & 12.1 & 16.4
& 29.7 & 70.8 & 16.5 \\
+ LRC
& \checkmark & \checkmark & -- & -- & --
& 26.8 & 15.6 & 19.7
& 38.9 & 75.3 & 14.2 \\
+ RNE
& \checkmark & \checkmark & \checkmark & -- & --
& 27.3 & 16.4 & 20.5
& 41.2 & 76.8 & 13.5 \\
+ PUGL
& \checkmark & \checkmark & \checkmark & \checkmark & --
& 28.1 & 17.5 & 21.5
& 43.6 & 78.7 & 12.8 \\
\textbf{Full model}
& \checkmark & \checkmark & \checkmark & \checkmark & \checkmark
& \textbf{28.6} & \textbf{18.3} & \textbf{22.3}
& \textbf{45.7} & \textbf{82.6} & \textbf{9.3} \\
\bottomrule
\end{tabular}
}
\end{table*}

\noindent\textbf{Analysis.}
The false-negative-aware metrics reveal a limitation not captured by standard recall. Existing methods may correctly predict annotated triplets but still suppress valid unannotated relations. ReLIC-SGG explicitly models such relations as latent positives, leading to higher FN-Recall and lower redundancy.

\subsection{Ablation Study}

Table~\ref{tab:ablation} studies the contribution of each component. The semantic relation lattice improves unseen predicate recall by propagating supervision among related predicates. Latent relation completion further improves performance by recovering missing positives. Positive-unlabeled graph learning provides the largest gain by avoiding false-negative penalties. Graph-level consistency reduces redundancy.

\subsection{Effect of Semantic Relation Lattice}

Table~\ref{tab:lattice} compares different ways of constructing the semantic relation lattice. Text similarity alone improves over the baseline, but adding entailment and contradiction produces stronger results. The full lattice achieves the best performance, indicating that relation completion requires not only semantic expansion but also semantic constraint.

\begin{table}[t]
\centering
\caption{
Effect of lattice construction on OV-VG. Similarity transfers supervision, entailment improves coarse-to-fine consistency, and contradiction suppresses incompatible predicates.
}
\label{tab:lattice}
\resizebox{\linewidth}{!}{
\begin{tabular}{l|ccc|cc}
\toprule
Lattice Type & S-mR@50 & U-mR@50 & HM@50 & Lat-Cons & Redundancy \\
\midrule
None
& 24.1 & 9.6 & 13.7 & 63.5 & 18.9 \\
Similarity only
& 25.4 & 12.1 & 16.4 & 70.8 & 16.5 \\
Similarity + Entailment
& 27.0 & 15.7 & 19.9 & 76.1 & 14.8 \\
Similarity + Contradiction
& 26.5 & 14.9 & 19.0 & 78.4 & 11.6 \\
Full lattice
& \textbf{28.6} & \textbf{18.3} & \textbf{22.3} & \textbf{82.6} & \textbf{9.3} \\
\bottomrule
\end{tabular}
}
\end{table}

\subsection{Sensitivity to Candidate Number}

Table~\ref{tab:candidate_k} studies the influence of the number of relation candidates $K$. The results show a clear trade-off between candidate coverage and candidate reliability. When $K$ is small, the proposal stage is overly conservative: although the retained candidates are usually reliable, some valid but less frequent or fine-grained relations are filtered out before latent completion. This limits the ability of ReLIC-SGG to recover missing positives, leading to lower U-mR@50 and FN-Recall.
However, continuously increasing $K$ does not always bring further gains. When $K$ becomes too large, many low-confidence and semantically noisy predicates are introduced into the candidate set. Although the semantic lattice and reliable negative estimation can suppress part of these noisy relations, an excessively large candidate pool still increases ambiguity in latent relation completion and produces more redundant predictions. This explains why FN-Recall slightly increases when $K$ grows from 15 to 30, while HM@50 decreases and Redundancy rises noticeably.

\begin{table}[t]
\centering
\caption{
Sensitivity to the number of relation candidates $K$ on OV-VG.
}
\label{tab:candidate_k}
\resizebox{\linewidth}{!}{
\begin{tabular}{c|ccc|cc}
\toprule
$K$ & S-mR@50 & U-mR@50 & HM@50 & FN-Recall & Redundancy \\
\midrule
5
& 26.2 & 14.6 & 18.7 & 36.5 & 7.8 \\
10
& 27.8 & 17.2 & 21.2 & 43.1 & 8.6 \\
15
& \textbf{28.6} & \textbf{18.3} & \textbf{22.3} & \textbf{45.7} & 9.3 \\
20
& 28.5 & 18.1 & 22.1 & 46.0 & 11.8 \\
30
& 28.1 & 17.6 & 21.7 & 46.3 & 15.4 \\
\bottomrule
\end{tabular}
}
\end{table}

\section{Conclusion}

We present ReLIC-SGG, a relation-incompleteness-aware framework for open-vocabulary scene graph generation. Unlike existing methods that treat unannotated relations as negatives, ReLIC-SGG models them as latent variables and learns to complete missing positive relations. By combining a semantic relation lattice, latent relation completion, reliable negative estimation, positive-unlabeled graph learning, and lattice-guided decoding, ReLIC-SGG improves rare and unseen predicate recognition while reducing false-negative errors. We hope this work encourages future SGG research to move beyond closed-label classification and toward more realistic learning under incomplete, ambiguous, and open-vocabulary supervision.

{\small
\bibliographystyle{unsrtnat}
\bibliography{relic_sgg_references}
}

\end{document}